\def\year{2022}\relax
\documentclass[letterpaper]{article} 
\usepackage{aaai22}  
\usepackage{times}  
\usepackage{helvet}  
\usepackage{courier}  
\usepackage[hyphens]{url}  
\usepackage{graphicx} 
\usepackage{natbib}  
\usepackage{caption} 
\DeclareCaptionStyle{ruled}{labelfont=normalfont,labelsep=colon,strut=off} 
\usepackage{algorithm}
\usepackage{algorithmic}

\usepackage{microtype}
\usepackage{subdepth}
\usepackage{amsmath}
\usepackage{multirow}

\usepackage{booktabs}

\usepackage{amsmath}
\usepackage{amssymb}

\newcommand{\R}{\mathbb{R}}

\usepackage{newfloat}
\usepackage{listings}
\floatstyle{ruled}
\newfloat{listing}{tb}{lst}{}
\floatname{listing}{Listing}

\begin{document}

\title{KARL-Trans-NER: Knowledge Aware Representation Learning for Named Entity Recognition using Transformers}
\author{

    Avi Chawla,
    Nidhi Mulay,
    Vikas Bishnoi,
    Gaurav Dhama
    
}
\affiliations{

    {Mastercard AI, Gurgaon, India} \\
\{avi.chawla, nidhi.mulay, vikas.bishnoi, gaurav.dhama\}@mastercard.com

}

\maketitle

\begin{abstract}

The inception of modeling contextual information using models such as BERT, ELMo, and Flair has significantly improved representation learning for words. It has also given SOTA results in almost every NLP task — Machine Translation, Text Summarization and Named Entity Recognition, to name a few. In this work, in addition to using these dominant context-aware representations, we propose a \underline{K}nowledge \underline{A}ware \underline{R}epresentation \underline{L}earning (KARL) Network for Named Entity Recognition (NER). We discuss the challenges of using existing methods in incorporating world knowledge for NER and show how our proposed methods could be leveraged to overcome those challenges. KARL is based on a Transformer Encoder that utilizes large knowledge bases represented as fact triplets, converts them to a graph context, and extracts essential entity information residing inside to generate contextualized triplet representation for feature augmentation. Experimental results show that the augmentation done using KARL can considerably boost the performance of our NER system and achieve significantly better results than existing approaches in the literature on three publicly available NER datasets, namely CoNLL 2003, CoNLL++, and OntoNotes v5. We also observe better generalization and application to a real-world setting from KARL on unseen entities.


\end{abstract}

\section{Introduction}

Named Entity Recognition (NER) is the task of locating and classifying named entities in a given piece of text into pre-defined entity categories such as Person (PER), Location (LOC), Organisation (ORG), etc. NER is considered an essential preprocessing step that can benefit many downstream applications in Natural Language Processing (NLP), such as Machine Translation \cite{babych-hartley-2003-improving}, Information Retrieval \citep{articleir_ner} and Text Classification \citep{article_tc_ner}.

Over the past few years, Deep Learning has been the key to solving not only NER but many other NLP applications \citep{le-etal-2018-deep, kouris-etal-2019-abstractive}. 
On the downside, these models also demand a lot of well-structured and annotated data for their training. This restricts the applicability of trained models to a real-world scenario as the model's behavior and predictions become very specific to the type of data they are trained on. To conquer this, many studies have recently evolved that focus on building models that can incorporate world knowledge for enhanced modeling and inference on the task at hand, such as \citet{qizhen2020} for NER, \citet{denk-peleteiro-ramallo-2020-contextual} for Representation Learning and \citet{7451560} for Dependency Parsing, etc. 

Although recent works in literature have successfully incorporated world knowledge for Sequence Labeling \citep{qizhen2020}, they come with certain limitations, which we discuss ahead. First, as words in a language can be polysemous \citep{The_Structure_of_Polysemy}, entities and relations in a knowledge graph can be polysemous too \citep{xiao2015}. To introduce Knowledge Graph Embeddings (KGEs), we noticed that previously proposed approaches have primarily used pre-trained static embeddings obtained from extensive sources such as Wikidata. KGEs in these models fundamentally relies on the assumption that the tail entity is a linear transformation of the head entity and the relation, making them non-contextualized in nature. Second, we noticed that prior work only considered head-entity and relation embedding to get the knowledge graph embedding and ignored the tail-entity of the triplet completely. Dropping the tail entity entirely could lead to a potential loss of information. We observed that in addition to carrying information about the triplet itself, the head-relation-tail also helps in understanding and extracting implicit relationships existing between entities across triplets. Therefore, the model must know where the head and the relation are leaning towards to achieve accurate embedding estimation. The final limitation lies in applying a Recurrent architecture to obtain KGEs, introducing time inefficiency and a high computation cost \citep{annervaz2018learning}. 


To further understand the importance and our motivation behind using world knowledge for NER, consider a couple of examples mentioned below.\\

\noindent A: \emph{Google announced the launch of Maps.}\\
B: \emph{Pichai announced the launch of Maps.}\\
In the training phase, the significant contextual overlap between sentences can confuse the model in labeling the named entities correctly. The model is likely to memorize sentence templates rather than learn to predict correct entity labels, leading to misclassifications. Also, suppose we trained our NER model on an out-of-domain NER dataset. In that case, the model would have hardly received any information from the training set that "Google" is an organization and "Pichai" indicates a Person. \\

\noindent A: \emph{Berlin died in Season 2 of Money Heist.}\\
B: \emph{Messi saved Barcelona with an equalizer.}\\
In the first sentence above, "Berlin" refers to a person, whereas in the second sentence, "Barcelona" refers to an organization. There are reasonable chances of misclassification in these two sentences because of a high probability of training data missing such nuance differences in all the possible entity tags for a named entity.

From the examples mentioned above, we can infer that for the model to be aware of such subtle differences, we should provide it with the ability to look up relevant details from a reliable source. Therefore, world knowledge can open the gates for the model to access such information and learn details about entities that it might never come across in the training data. In addition to this, with access to structured world knowledge, far better applicability to a real-world setting can be expected. 

Setting these points as our objective, in this work, we propose Knowledge Aware Representational Learning Network for Named Entity Recognition using Transformer (KARL-Trans-NER), which 
\begin{enumerate}
    
    \item Encodes the entities and relations existing in a knowledge base using a self-attention network to obtain Knowledge Graph Embeddings (KGEs). The embeddings thus obtained are dynamic and fully contextualized in nature.
    \item Takes the encoded contextualized representations for entities and relations and generates a knowledge-aware representation for words. The representation obtained, which we also call "Global Representation" for words, can be augmented with the other underlying features to boost the NER model's performance. 
    \item Generates sentence embeddings using BERT by fusing task-specific information through NER tag embeddings.
    \item And lastly, relies on a Transformer as its context encoder incorporating direction-aware, distance-aware, and un-scaled attention for enhanced encoder representation learning. 
\end{enumerate}

To verify the effectiveness of our proposed model, we conduct our experiments on three publicly available datasets for NER. These are CoNLL 2003 \citep{DBLP:journals/corr/cs-CL-0306050}, CoNLL++ \citep{wang2019crossweigh} and OntoNotes v5 \citep{pradhan-etal-2013-towards}. Experimental results show that the global embeddings generated for every word using KARL, when used for feature augmentation, can result in significant performance gains of over 0.35-0.5 \emph{F}\textsubscript{1} on all the three NER datasets. Also, to validate the model's generalizability and applicability in a real-world setting, we generate the model's prediction on random texts taken from the web. Results suggest that incorporating world knowledge enables the model to make accurate predictions for every entity in the sentence. 

\section{Related work}
The research community in NER moved from approaches using character and word representations \citep{yao2015, zhou2017, kuru2016} to sentence-level contextual representations \citep{yang2017, zhang2018}, and recently to document-level representations as proposed by \citet{qian2018} and \citet{akbik2018}. Expanding the scope of embeddings from character and word level to document level has shown significant improvements in the results for many NLP tasks, including NER \citep{luo2019hierarchical}. To expand the scope further, researchers have explored external knowledge bases to learn facts existing in the universe that may not be present in the training data \citep{annervaz2018learning, qizhen2020}.

Incorporating information present in Knowledge Graph is an emerging research topic in NLP.
While some methods focus on graph structure encoding \citep{lin-etal-2015-modeling, das-etal-2017-chains}, others focus on learning entity-relation embeddings \citep{wang2020coke, jiang-etal-2019-adaptive}. 

\citet{zhong2015} proposed an alignment model for jointly embedding a knowledge base and a text corpus that achieved better or comparable performance on four NLP tasks: link prediction, triplet classification, relational fact extraction, and analogical reasoning.
\citet{xiao2015} proposed a generative embedding model, TransG, which can discover the latent semantics of a relation and leverage a mixture of related components for generating embedding. They also reported substantial improvements over the state-of-the-art baselines on the task of link prediction. 
\citet{10.1145/3038912.3052675} presented a neural network to answer simple questions over large-scale knowledge graphs using a hierarchical word and character-level question encoder.
\citet{annervaz2018learning} leveraged world knowledge in training task-specific models and proposes a novel convolution-based architecture to reduce the attention space over entities and relations. It outperformed other models on text classification and natural language inference tasks.

Despite producing state-of-the-art results in many NLP tasks, Knowledge Graphs are relatively unexplored for NER. \citet{qizhen2020} introduced a Knowledge-Graph Augmented Word Representation (KAWR). The proposed model encoded the prior knowledge of entities from an external knowledge base into the representation. Though KAWR performed better than its benchmark BERT \citep{bert}, the model underperformed compared to the SOTA models for NER. 

\section{Knowledge Augmented Representation Learning Network}

In this section, we describe the end-to-end proposed model for the task of NER (KARL-Trans-NER). 

\subsection{Knowledge Graph Embedding Model}

World Knowledge is represented in the form of fact triplets in a Knowledge Graph, such as \emph{("Albert Einstein", "BornYear", "1879")}. Like any other representation technique, to incorporate the information residing inside these fact triples, they need to be encoded into a numeral representation.
To take into account polysemy and learn graph embeddings as a function of the graph context, we take inspiration from CoKE \citep{wang2020coke} and train our knowledge graph embedding model on an entity prediction task using the idea of Masked Language Modeling \citep{bert}. \\

\noindent \textbf{Model Architecture}

\noindent Our KGE model is based on a Transformer architecture \citep{vaswani2017attention}. The corresponding architecture is shown in Figure 1. Training objective of our KG embedding module is inspired from that of CoKE. Since its application to a task like NER was not straightforward, we designed a character-level rich knowledge graph embeddings to understand the underlying representation of characters in a KG. For, e.g., in the fact-triplet \emph{"(BarackObama, HasChild, SashaObama)"}, the tokens do not appear the way words do in English sentences. Introducing a character-level representation layer in CoKE can help the model understand the intrinsic character sequence representations of entities and relations alongside word-sequence representation to achieve an enhanced representation learning of the knowledge graph. Also, as a collection of fact triplets in the form of a knowledge graph is an ever-expanding resource, only entity/relation level embeddings will not adapt to newer entities, thereby introducing trouble in obtaining embeddings that were not observed during training the KGEs. We used two different character sequence encoders, one for the entities and another for the relations. Also, instead of feeding sinusoidal positional embeddings with the input embeddings \citep{vaswani2017attention}, we adopt relative positional embeddings \citep{shaw-etal-2018-self, DBLP:journals/corr/abs-1901-02860}. Positional embeddings are distance-aware, but they are unaware of the directionality. Our intuition of using relative positional embeddings here relies on the findings of \citet{yan2019tener}.\\

\noindent \textbf{Training Data Preparation}

\noindent We are given a Knowledge Graph composed of fact triplets as follows:

\begin{center}

$ KG = \{<s,r,o> | (s,o) \in E, r \in R\} $
\end{center}

\noindent where, $s, o$ is the subject and object entity respectively, $r$ is the relation between them, $E$ is the entity set and $R$ is the relation set. A sequence of tokens or a context is created using each triplet as $\emph{s} \rightarrow \emph{r} \rightarrow \emph{o}$. 
All the triplets in the knowledge graph are formulated in this manner to obtain a set of graph contexts as follows:

\begin{center}
$ S = \{(s \rightarrow r \rightarrow o) | (s,o) \in E, r \in R\} $
\end{center}
\noindent Next, we create the training set $T$ from $S$. This is done by replacing each object entity $o$ in $S$ with the "[MASK]" token and defining the object entity $o$ as the prediction label for the triplet as follows: 
\begin{center}
$ T = \{(<s,r,[MASK]>, o)\} $

\end{center}

\noindent \textbf{Model Training}

\begin{figure}[t]
\includegraphics[width=0.49\textwidth]{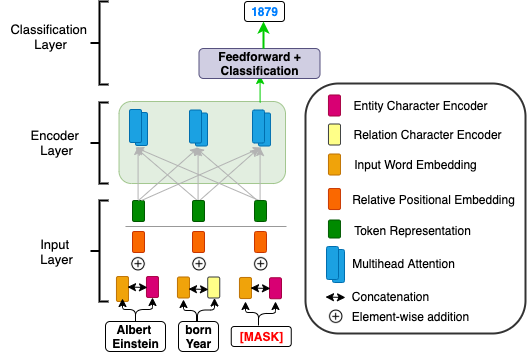}
\caption{The overall architecture of the Knowledge Graph Embedding Learning Network.}
\end{figure}

\noindent Consider an input sequence $T$ = $(t_{1}, t_{2}, t_{3})$. 
First, each token $t_{i}$ is passed through its corresponding character level encoder to obtain the character sequence representation. Further, for each token $t_{i}$ in the input sequence $T$, we obtain its word-level representation which is tuned during model's training. The character sequence representation  $(x_{i}^{char})$ and the word-level representation $(x_{i}^{word})$ are then concatenated together to obtain the element or the token embedding $(x_{i}^{ele})$ for $t_{i}$.

$$x_{i}^{ele} = [x_{i}^{word} ; x_{i}^{char}]$$


The final embedding input $(h_{\emph{\text{i}}}^{\text{0}})$ which is given as an input to the transformer encoder for token $t_{i}$ is obtained by the element wise sum of its element embedding $(x_{i}^{ele})$ and its relative positional embedding $(x_{i}^{pos})$. 

\begin{center}
$h_{i}^{0} = x_{i}^{ele} \oplus x_{i}^{pos}$ 
\end{center}

Once the input representation is generated, it is fed to a transformer encoder with $L$ successive layers. The hidden state for token $t_{i}$ at layer $j$ is denoted as $h_{i}^{j}$ and is given by:

\begin{center}
$h_{i}^{j}$ = TransEnc($h_{i}^{j-1}$), $j = 1,2, \dots, L$
\end{center}

We treat the hidden representations obtained at the very last layer as the output of the transformer encoder. This is denoted by $\{h_{1}^{L}, h_{2}^{L}, h_{3}^{L}\}$. 

After obtaining the transformer encoder representations of the last layer, $\{h_{i}^{L}\}_{i=1}^{3}$, we select the encoder representation corresponding to the "[MASK]" token, i.e., $h_{3}^{L}$. This is fed through a feedforward layer, which is followed by softmax classification layer to predict the third token, or the object entity $t_{3}$ in $(t_{1}, t_{2}, t_{3})$.
Mathematically, the above feedforward layer and softmax classification layer is defined as follows:

\begin{center}
$f$ = $W*h_{3}^{L} + b$ 
\end{center}

\begin{center}
$o$ = $\frac{exp(f_{k})}{\sum_{k}^{}exp(f_{k}))}$
\end{center}

Here, $W \in \R^{V\times D}$ and $b \in \R^{V\times 1}$ are learnable parameters of the feedforward layer, $V$ is entity vocabulary size, $D$ is the hidden size, $f$ is the output of the feedforward layer and $o$ is the predicted probabilities through softmax layer over all the entities in the entity vocabulary.

The model is trained using Adam Optimizer \citep{kingma2017adam} and we define training loss as the cross-entropy loss \citep{10.5555/646815.708603} between the predicted entity probabilities $o$ and one-hot ground truth entity $p$ as follows:
\begin{center}
$loss$ = $ - \sum_{k}^{}p_{k}\log{o_{k}}$ 
\end{center}
Here, $p_{k}$ and $o_{k}$ are the $k^{th}$ components of $p$ and $o$ respectively. 


\subsection{NER Model}

The architecture of the proposed model is shown in Figure 2. We describe each component of our NER model in detail below.

\subsubsection{Input Representation Layer}

Given an input sequence of tokens $S = (x_{1}, x_{2}, \dots, x_{n})$, we first obtain the embeddings by generating features in six different ways of varying scope. These features are defined below:\\

\noindent \textbf{Word-level Representations:} For word-level features, we use the pre-trained 100-dimensional Glove Embeddings \citep{pennington-etal-2014-glove} and tune them during model's training. \\

\noindent \textbf{Character-level Representations:} Using word-level representations alone typically is not considered the best approach to NER \citep{santos2015boosting} due to the out-of-vocabulary (OOV) problem. To address this, many neural NER systems have shown the effectiveness of incorporating character-level representations for words \citep{ma2016endtoend, chiu-nichols-2016-named}. In addition to solving the OOV problem, they also help the model understand the underlying structure of words. This includes learning the arrangement of chars, the distinctive features a named entity follows such as the capitalization of the first letter of named entities, etc. Most works in the literature prefer a CNN-based character-level encoder over an LSTM based encoder \citep{li-etal-2017-leveraging-linguistic}. This is because of CNN's parallelization capabilities and almost similar or even better performance than the latter. To enrich our NER model with character-level features, we adopt IntNet \citep{intnet}, a funnel-shaped convolutional neural network. IntNet combines kernels of various sizes to extract different n-grams from words and learn an enhanced representation of words. 

The entire network of IntNet comprises a series of convolution blocks, and each block consists of two layers. The first layer is a basic $N \times 1$  convolution on the input. The second layer applies convolutions of different kernel sizes on the first layer's output, concatenates them, and feeds it to the next convolution block.\\

\begin{figure}
\centering
\includegraphics[width=0.40\textwidth, height = 6cm]{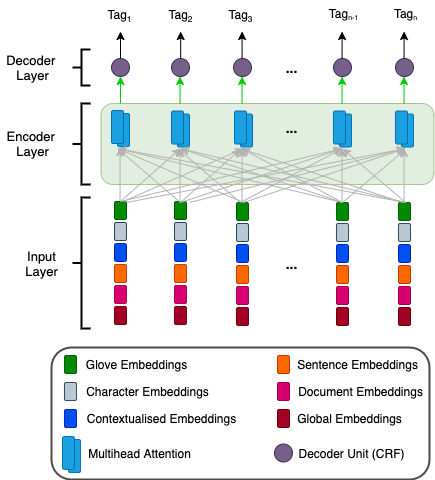}
\caption{The detailed architecure of the proposed KARL-Trans-NER model.}
\end{figure}


\noindent \textbf{Context-level Representations:} To enlighten the model with contextual knowledge, here, we use the dominant pre-trained contextualized embedding model BERT \citep{bert}. As BERT generates embeddings at the word-piece-level, we take the average of all the word pieces to obtain the contextualized word embedding of a word. \\

\noindent \textbf{Sentence-level Representations:} Most sentence embedding learning techniques rely on word-level embeddings to generate sentence-level features \citep{8639211}. For computational ease, some methods compute the average of all the token embeddings in a sentence \citep{annervaz2018learning, coates-bollegala-2018-frustratingly} to obtain the sentence representation.
\begin{center}
    $s$ = $\frac{\sum_{i} w_{i}}{n}$ 
\end{center}
where, $n$ in the sequence length and $w_{i}$ is the word embedding of token $i$. 
Averaging words embeddings neglects the word order information. It also assumes an equal contribution from each word towards the sentence representation. Therefore, it is not the best approach to obtain sentence embeddings. 

Sentence transformers proposed by \citet{reimers2019sentencebert} generate sentence embeddings using contextualized word embeddings, making them adaptive to new contexts. Though sentence transformers have provided SOTA results in many NLP tasks \citep{ke2020sentilare, Li_2020}, we believe that these representations could be enhanced by incorporating task-specific information. To achieve this, we adopt the techniques proposed by \citet{wang2018joint} and incorporate task-specific information by learning the NER label embeddings. We discuss our approach of generating sentence embeddings using the contextualized word embeddings obtained from BERT in detail below.

We begin by first taking the contextualized representations of words obtained from BERT and linearly projecting them to a different space. Let us denote the obtained representations by $c = (c_{1}, c_{2},\dots, c_{n})$, where $c \in \R^{n\times d}$, $c_{i}$ denotes the contextualized embedding of word $i$, $n$ is the number of words in the sentence and $d$ is the dimension BERT embeddings were projected to. The next step involves embedding the labels (PER, LOC etc.) to the same space. The unique set of labels in the data are embedded into dense representations. Let these representations be defined as $l = (l_{1}, l_{2}, \dots, l_{m})$. Here, $l \in \R^{m\times d}$, $m$ denotes the number of tag labels and $d$ is the hidden size. 

Next, we apply the label embedding attention over the word representations. Cosine-similarity between each token embeddings $c_{i}$ and label embedding $l_{j}$ is treated as the compatibility function. 

\begin{center}
    $simi(c_{i}, l_{j}) = \frac{c_{i}^{T}l_{j}}{||c_{i}|| ||l_{j}||}$
\end{center}


    


A convolution operation is applied to the similarities obtained above. This enables the model to capture the behavior of the same label in its neighboring words. In other words, the application of a convolution layer captures the relative spatial information for a particular label over a phrase. 

Considering a phrase of length $2k+1$ where $k$ is the kernel size, the convolution for token $i$ is applied over $simi_{[i-k:i+k, :]}$ which is followed by max-pooling. The attention weights for the entire sentence are generated after that by applying softmax operation over the scores. Finally, sentence-level representation $s$ is computed as the weighted sum of the token embeddings over the attention weights. This is mathematically defined as follows: 
\begin{center}
    $scores = max(W^{T} simi_{[i-k:i+k, :]} + b) $ \\
    
    $\alpha = softmax(scores)$
    
    $s = \sum_{i=1}^{n} \alpha_{i}c_{i}$ \\
\end{center}
Here, $W \in \R^{2k+1}$ and $b \in \R^{m}$ are learnable parameters, $\alpha \in \R^{n}$ and $s \in \R^{d}$.\\

\noindent \textbf{Document-level Representations:} To obtain Document-level features, we adopt a key-value Memory Network \citep{weston2015memory} that generates document-aware representations of every unique word in the training data. Specifically, we refer to the model adopted by \citet{luo2019hierarchical} to create features at a document level. \\

\noindent \textbf{Global-level Representations:} 
\citet{annervaz2018learning} was among the first attempts to augment learning models with structured graph knowledge. They tested their model in a sentence classification setting. To retrieve entities and relations from the knowledge base, they first clustered entities and relations. A convolution network was used to obtain the cluster representation. Though the technique did work well for sentence classification, we identified a couple of limitations in their approach.

The first lies in the clustering step. To retrieve relevant entities and relations, clustering should be accurate. Also, clustering generally results in a loss of information. The cluster representations are always assumed to be a generalized representation of its constituents. Being generalized, it can neglect essential information existing inside the cluster's objects. Second, they augmented their model with graph knowledge at a sentence level. For sequence labeling, augmentation at the sentence level will not be effective as we require precise entity-level information from the knowledge graph. We address the limitations mentioned above and propose more efficient and reliable technique for incorporating relevant information from the knowledge graph. 

Instead of performing clustering, we first shortlist entities and relations that are relevant to our task. The shortlisted entities and relations are passed through the already trained knowledge graph embedding module. Here, we remove the classification layer of the knowledge graph embedding module and keep the representations generated by the transformer network to create the fact triplet embeddings. Next, we discuss the entity and relation shortlisting step. \\

\noindent \textbf{Entity Shortlisting:} One caveat of similarity-based methods is that their performance heavily depends on the modeling of adopted features for similarity estimation. This can introduce errors. Therefore, we adopt an n-gram based matching with the input document to shortlist entities from the Knowledge Graph. The idea here is to generate a candidate set for each entity in the document. More specifically, we refer to the rules proposed by \citet{10.1145/3038912.3052675} for candidate set generation. As a particular entity in the document can match with multiple entities in the knowledge graph, we first rank all the entities based on the number of triplets they appear as subjects in the knowledge graph. The top $k_{1}$ entities with the highest rank are selected and added to the candidate entity set $C_{w}$ of the current word $w$. \\

\noindent \textbf{Relation Shortlisting:} After generating the candidate entity set $C_{w}$ for a word $w$, we extract the $k_{2}$ most frequent relations that emerge from entities in the candidate entity set $C_{w}$. This results in the Entity Relation set $ER_{w}$ for a word $w$ as follows:
\begin{equation*}
  ER_{w} = \Bigg\{
  \begin{matrix}
    (e_{1}, r_{1}) & (e_{1}, r_{2}) & \dots & (e_{1}, r_{k_{2}})\\
    (e_{2}, r_{1}) & (e_{2}, r_{2}) & \dots & (e_{2}, r_{k_{2}})\\
    
    (e_{k_{1}}, r_{1}) & (e_{k_{1}}, r_{2}) & \dots & (e_{k_{1}}, r_{k_{2}})
\end{matrix}
   \Bigg\}
\end{equation*}

After shortlisting relevant entities and relations, we augment the current word with world knowledge. We do this by first encoding all the triples in the set $ER_{w}$ using the pre-trained knowledge graph embedding model. Each entity-relation pair in $ER_{w}$ is appended with the [MASK] token and fed to the transformer network. The hidden state obtained from the last layer of the transformer network acts as the contextualized representation for these triplets. Finally, we apply soft-attention over the hidden states obtained from the transformer encoder and treat the input contextualized embeddings obtained from BERT as the query vector for augmentation at word-level. For a word $w$, this is mathematically defined as follows:

\begin{center}
    $I = [h_{w,1}^{L}; h_{w,2}^{L}; h_{w,3}^{L}] $\\
    
    $Q, K, V = W^{q}B_{w}, W^{k}I, W^{v}I $\\
    
    $A_{j} = QK_{j}^{T}$\\
    
    $g_{w} = \sum_{i}Softmax(\frac{A_{i}}{\sqrt 3d})V_{i}$

\end{center}

Here, $W^{q} \in \R^{3d\times h}$, $W^{k} \in \R^{n \times 3d}$ and $W^{v} \in \R^{n \times 3d}$ are learnable parameters, $B_{w} \in \R^{h}$ is the contextualized word embedding of word $w$ obtained from BERT, $I \in \R^{n\times3d}$ is the concatenation of the individual hidden states $\{h_{w,i}^{L}\}_{i=1}^{3}$ corresponding to the last layer of the transformer encoder, $n$ is the number of triplets in the set $ER_{w}$, $d$ is the output dim of individual hidden state from the transformer encoder and lastly, $g_{w} \in \R^{3d}$ is the global representation of word $w$.

This concludes our input representation layer. The output of the input representation layer is treated as the concatenation of the above-described features at token-level to obtain the word-level representation, which is 1) character-aware, 2) word-aware, 3) context-aware, 4) sentence-aware, 5) document-aware, and 6) knowledge-aware. 

\subsubsection{Encoder Layer}

As a context-encoder, we use the fully-connected self-attention Network, aka Transformer network. For NER, \citet{guo2019startransformer} found the transformer encoder to perform poorly compared to a recurrent context encoder like LSTM \citep{articlelstm}. \citet{yan2019tener} proposed that the use of un-scaled attention and relative positional encodings in place of positional encoding can significantly improve the performance of the Transformer. Thus, we also adopt their proposed modifications in the Transformer based context encoder. 

\subsubsection{Decoder Layer}
Conditional random field (CRF) \citep{10.5555/645530.655813} is a widely adopted decoder in many state-of-the-art NER models \citep{ma2016endtoend, lample-etal-2016-neural}. CRF can establish strong connections between the output tags, which help in making better predictions over the softmax layer. Viterbi Algorithm is applied in the decoding phase to obtain the final output label sequence with the highest probability out of all the valid label sequences. 

\section{Experiments and Results}

We evaluated our models on three NER tasks. To reduce the impact of randomness, we conducted each experiment three times and reported the average span-level F1 score and standard deviation. Starting with word-level, we incrementally conducted our experiments and augmented features in increasing order of the extent of information modeled by them. To verify the performance of these trained models in a real-world scenario and compare their adaptability to unseen entities, we also generated predictions on two random pieces of texts taken from the web. In the subsequent section, we discuss the data statistics, model details, results, and performance on unseen entities.\\
\begin{table}[t]
\resizebox{0.49\textwidth}{!}
{%

\begin{tabular}{r|c|c|c|c|c}
\toprule
\textbf{Dataset} & \textbf{Type} & \textbf{Train} & \textbf{Test} & \textbf{Dev} & \textbf{Tags} \\
\midrule
\multirow{2}{*}{CoNLL 2003} & Sentence & 14041 & 3453 & 3250 & \multirow{2}{*}{4}\\
& Token & 203621 & 46435 & 51362 \\
\midrule

\multirow{2}{*}{OntoNotes v5} & Sentence & 59924 & 8262 & 8528 & \multirow{2}{*}{18}\\
& Token & 1088503 & 152728 & 147724 \\

\bottomrule
\end{tabular}%
}
\caption{The table represents the details for CoNLL2003 and OntoNotes v5 datasets. The statistics for CoNLL++ dataset are same as that of CoNLL 2003.}
\end{table}

\noindent \textbf{NER datasets and Knowledge Graph}

\noindent The statistics of the three NER datasets we experimented on are listed in Table 1. Wikidata is one of the largest sources of real-world knowledge data covering concepts belonging to various domains. We filter the Wikidata and use only those fact triplets that were relevant to our task. This left us with approximately 10 million fact triplets out of roughly 400 million fact triplets present in Wikidata.\\


\noindent \textbf{Model Hyperparameters}

\noindent We tuned all the model hyper-parameters manually, and list down the respective search ranges for all the hyper-parameters involved in Table 2. \\

\begin{table}[t]
\resizebox{0.49\textwidth}{!}
{%

\begin{tabular}{p{0.20\textwidth} p{0.20\textwidth}||p{0.20\textwidth}p{0.20\textwidth}}
\toprule
\multicolumn{2}{c||}{\textbf{NER-Transformer}} & \multicolumn{2}{c}{\textbf{Knowledge-Graph Transformer}} \\
\midrule
Layers &  ~\hfill~[2] ~\hfill~ & Layers &  ~\hfill~[2] ~\hfill~ \\
\midrule
Learning Rate &   ~\hfill~[0.001, 0.0009] ~\hfill~ & Learning Rate &   ~\hfill~[0.0005, 0.0003]  ~\hfill~ \\
\midrule
Heads &   ~\hfill~[8, 12, 14]  ~\hfill~ & Heads &   ~\hfill~[4, 8]  ~\hfill~\\
\midrule
Head Dim. &   ~\hfill~ [64, 96, 128]  ~\hfill~ & Head Dim. & ~\hfill~ [256, 128] ~\hfill~\\
\midrule
FC Dropout & ~\hfill~ [0.40] ~\hfill~  & FC Dropout & ~\hfill~ [0.40] ~\hfill~ \\
\midrule
Attn. Dropout & ~\hfill~ [0.15] ~\hfill~  & Attn. Dropout &  ~\hfill~ [0.25] ~\hfill~ \\
\midrule
Optimizer & ~\hfill~ [SGD] ~\hfill~  & Optimizer &  ~\hfill~ [Adam] ~\hfill~ \\
\midrule
\midrule
\multicolumn{4}{c}{\textbf{Other Hyperparameters and Model Settings}}\\
\midrule
\multicolumn{1}{c}{IntNet Layers} & \multicolumn{1}{c|}{[5]} & 
\multicolumn{1}{c}{IntNet Kernel Sizes} & \multicolumn{1}{c}{[[3,4,5]]}\\
\midrule
\multicolumn{1}{c}{IntNet Embedding Dim.} & \multicolumn{1}{c|}{[16, 32, 64]} &
\multicolumn{1}{c}{IntNet Hidden Dim.} & \multicolumn{1}{c}{[8, 16, 32]}\\

\midrule
\multicolumn{1}{c}{Momentum} & \multicolumn{1}{c|}{[0.9]} & \multicolumn{1}{c}{Epochs} & \multicolumn{1}{c}{[100]} \\

\midrule
\multicolumn{1}{c}{Sentence Transformer} & \multicolumn{1}{c|}{[stsb-bert-large]} & \multicolumn{1}{c}{Bert pre-trained} & \multicolumn{1}{c}{[bert-large-cased]} \\
\midrule
\multicolumn{1}{c}{Glove Embedding} & \multicolumn{1}{c|}{[glove-en-100d]} &  & \\

\bottomrule
\end{tabular}%
}
\caption{The table represents the hyper-parameter search ranges for the Recurrent and Transformer Encoder, and some other hyperparameters.}\label{tab:hyperparams}
\end{table}

\noindent \textbf{Results and Analysis}

\noindent We report the results achieved by our models in Table 3. Starting with word-level features, we observe that as augmentation is done with different features, the performance on every dataset consistently increases. For CoNLL datasets, the accuracy lift generated by each feature augmentation step follows the following trend: \emph{Char $>$ Context $>$ Global $>$ Doc $>$ Sent}. The same trend for OntoNotes v5 is \emph{Context $>$ Char $>$ Global $>$ Doc $>$ Sent}. 

\begin{table}[h]
\resizebox{0.49\textwidth}{!}
{
\begin{tabular}{p{0.20\textwidth}||c|c|c|c|c|c}
\toprule

\multicolumn{1}{c||}{\textbf{Model}} & \multicolumn{2}{c|}{\textbf{CoNLL 2003}} & \multicolumn{2}{c|}{\textbf{CoNLL++}} & \multicolumn{2}{c}{\textbf{OntoNotes}}\\
\midrule
\multicolumn{1}{c||}{LUKE \citep{yamada2020luke}} & \multicolumn{2}{c|}{\underline{94.3}} & \multicolumn{2}{c|}{-} & \multicolumn{2}{c}{-}\\
\midrule
\multicolumn{1}{c||}{\multirow{2}{*}{CMV \citep{luoma2020exploring}}} & \multicolumn{2}{c|}{{93.74 (0.25)}{$\bot$}} & \multicolumn{2}{c|}{-} &\multicolumn{2}{c} {-}\\
\multicolumn{1}{c||}{} & \multicolumn{2}{c|}{93.44 (0.06)} & \multicolumn{2}{c|}{-} & \multicolumn{2}{c}{-}\\
\midrule
\multicolumn{1}{c||}{ACE \citep{wang2020automated}} & \multicolumn{2}{c|}{{93.6}{$\bot$}} & \multicolumn{2}{c|}{-} & \multicolumn{2}{c}{-}\\
\midrule
\multicolumn{1}{c||}{CL-KL \citep{wang2021improving}} & \multicolumn{2}{c|}{{93.56}{$\bot$}} & \multicolumn{2}{c|}{\underline{94.81}} & \multicolumn{2}{c}{-}\\

\midrule
\multicolumn{1}{c||}{CrossWeigh \citep{wang2019crossweigh}} & \multicolumn{2}{c|}{{93.43}{$\bot$}} & \multicolumn{2}{c|}{{94.28}{$\bot$}} & \multicolumn{2}{c}{-}\\

\midrule
\multicolumn{1}{c||}{BERT-MRC+DSC \citep{li2020dice}} & \multicolumn{2}{c|}{93.33 (0.29)} & \multicolumn{2}{c|}{-} & \multicolumn{2}{c}{\underline{{92.07 (0.96)}}{$\bot$}} \\
\midrule
\multicolumn{1}{c||}{Biaffine-NER \citep{yu2020named}} & \multicolumn{2}{c|}{{93.5} $\bot$ } & \multicolumn{2}{c|}{-} & \multicolumn{2}{c}{91.30} \\
\midrule
\multicolumn{1}{c||}{KAWR \citep{qizhen2020}} & \multicolumn{2}{c|}{91.80 (0.24)}  & \multicolumn{2}{c|}{-} & \multicolumn{2}{c}{-} \\

\midrule
\midrule
\multicolumn{1}{c||}{\textbf{KARL-Trans-NER}} & F1 & $\triangle$ & F1 & $\triangle$ & F1 & $\triangle$ \\
\midrule
\multicolumn{1}{c||}{Word-level} & 90.12 (0.02) & - & 90.90 (0.01) & - & 86.97 (0.01) & -\\
\multicolumn{1}{c||}{+ Char-level} & 91.60 (0.01) & 1.48 & 92.53 (0.01) & 1.63 & 88.40 (0.08) & 1.43 \\
\multicolumn{1}{c||}{+ Context-level} & 92.92 (0.02) & 1.32 & 93.77 (0.05) & 1.24 & 90.13 (0.02) & 1.73 \\
\multicolumn{1}{c||}{+ Sentence-level}  & 93.10 (0.07) & 0.18 & 93.90 (0.07) & 0.13 & 90.42 (0.03) & 0.29 \\
\multicolumn{1}{c||}{+ Document-level} & 93.38 (0.04) & 0.28 & 94.17 (0.03) & 0.27 & 90.91 (0.04) & 0.49 \\
\multicolumn{1}{c||}{+ Global-level} & \textbf{93.74 (0.05)} & 0.36 & \textbf{94.52 (0.06)} & 0.35 & \textbf{91.41 (0.06)} & 0.50\\
\midrule
\multicolumn{1}{c||}{Context-level + Global-level} & 92.44 (0.09) & - & 92.98 (0.11) & - & - & - \\
\bottomrule

\end{tabular}
}
\caption{F1 scores on CoNLL 2003, CoNLL++ and OntoNotes v5. Results marked with {$\bot$} used both Train and Dev sets for model training. The SOTA results are underlined and the best results of our model are marked in bold. Standard deviation is written in parenthesis. {$\triangle$} denotes change in F1 by feature addition.}
\end{table}

In addition to this, we also observe that with access to world knowledge, the proposed model achieves superior results than most previously proposed systems existing in the literature. To better evaluate the effectiveness of our approach, we also compare our results with KAWR \citep{qizhen2020}. The authors also leveraged world knowledge to solve NER. To the best of our knowledge, their work is the only work in the entire literature that utilized knowledge graph embeddings and applied them to the datasets we used in our work. We compare their reported results with ours. An approximate lift of \emph{2-F\textsubscript{1}} is observed here. 

As the results reported by KAWR only utilized Context-level and Knowledge-level embeddings, we conducted our experiments with KARL using the exact configuration of the embeddings. Experimental results show that KARL achieved far better performance as compared to KAWR, outperforming it by approximately \emph{0.65} units on F\textsubscript{1} on the CoNLL 2003 dataset, which indicates the effectiveness of our proposed model in incorporating world knowledge. \\

\noindent \textbf{Evaluation on Unseen Entities}

\begin{table}[]
\resizebox{0.49\textwidth}{!}
{%
\begin{tabular}{@{}r|p{0.28\textwidth}|c|p{0.25\textwidth}|p{0.25\textwidth}@{}}
\toprule
\textbf{S.No} & ~\hfill~\textbf{Text}~\hfill~ & ~\hfill~\textbf{Model} ~\hfill~ & ~\hfill~\textbf{Predictions} ~\hfill~ & ~\hfill~\textbf{Predictions (lowercased)} ~\hfill~ \\
\midrule

\multirow{11}{*}{1} & \multirow{6}{*}{\parbox{0.28\textwidth}{[SpaceX (\emph{ORG})] is an aerospace manufacturer and space transport services company headquartered in [California (\emph{LOC})].}} & Word & None & None\\
\cline{3-5}
&  & \multirow{2}{*}{+ Char} & SpaceX (PER) & \multirow{2}{*}{None} \\
&  &  & California (LOC) &\\
\cline{3-5}
&  & \multirow{2}{*}{+ Context} & SpaceX (ORG) & SpaceX (MISC) \\
&  &  & California (LOC) & California (LOC)\\
\cline{3-5}
&  & \multirow{2}{*}{+ Sent} & SpaceX (ORG) & SpaceX (MISC)\\
&  &  & California (LOC) & California (LOC) \\
\cline{3-5}
&  & \multirow{2}{*}{+ Doc} & SpaceX (ORG) & SpaceX (MISC)\\
&  &  & California (LOC) & California (LOC)\\
\cline{3-5}
&  & \multirow{2}{*}{+ Global} & SpaceX (ORG) & SpaceX (ORG)\\
&  &  & California (LOC) & California (LOC)\\

\midrule
\multirow{13}{*}{2} & \multirow{6}{*}{\parbox{0.28\textwidth}{[Liverpool (\emph{ORG})] suffered an upset first time home league defeat of the season, beaten 1 by a [Guy Whittingham (\emph{PER}]) goal for [Sheffield Wednesday (\emph{ORG})].}} & Glove & None & None \\ 
\cline{3-5}
&  & \multirow{1}{*}{+ Int} & Liverpool (\emph{ORG}) & None\\
\cline{3-5}

&  & \multirow{2}{*}{+ Bert} & Liverpool (\emph{ORG}) & Liverpool (\emph{PER}) \\
&  &  & Sheffield (\emph{PER}) & Sheffield (\emph{PER}) \\
\cline{3-5}
&  & \multirow{3}{*}{+ Sent} & Liverpool (\emph{ORG}) & Liverpool (\emph{PER})\\
&  &  & Sheffield (\emph{PER}) & Sheffield (\emph{PER}) \\
&  &  & Whittingham (\emph{PER}) & Whittingham (\emph{PER}) \\
\cline{3-5}
&  & \multirow{3}{*}{+ Doc} & Liverpool (\emph{ORG}) & Liverpool (\emph{PER})\\
&  &  & Sheffield (\emph{PER}) & Sheffield (\emph{PER})\\
&  &  & Whittingham (\emph{PER}) & Whittingham (\emph{PER})\\
\cline{3-5}
&  & \multirow{3}{*}{+ KG} & Liverpool (\emph{ORG}) & Liverpool (\emph{ORG})\\
&  &  & Sheffield Wednesday  (\emph{ORG}) & Sheffield Wednesday  (\emph{ORG})\\
&  &  & Guy Whittingham (\emph{PER}) & Guy Whittingham (\emph{PER})\\

\bottomrule
\end{tabular}%
}
\caption{Table shows the predictions made by each of the 6 models we trained. "Predictions" column lists the entities and the tags predicted by the model on the corresponding Text. "Predictions (lowercased)" indicates the predictions on the lowercased version of Text.}

\end{table}

\noindent Next, we test each model's performance on unseen entities. We take the models trained on the CoNLL 2003 dataset and predict named entities for random pieces of texts taken from the web, which we annotated manually. We show the entities identified and their corresponding entity tags as predicted by these models for two such sentences in the "Predictions" column of Table 4. We observe that the word-level model failed to identify any named entity from the first sentence. With the introduction of character-level features, the model made one classification error and classified "SpaceX" as a Person rather than an Organisation. After augmenting features at the context level, all the subsequent models generated accurate predictions and identified all the named entities and the corresponding tags precisely.

In the second sentence, the word-level model again failed to identify any named entities from the input text. Although the model did start to identify parts of named entities with further feature augmentations, we observe that majority of the tag labels predicted were wrong. For instance, the "+Bert", "+Sent" and "+Doc" models misclassified "Sheffield" as a person. It is not until the augmentation of world knowledge do we start observing accurate predictions. The Global model, just as before, achieved an accuracy of 100\% in named entity identification and tag-labeling. 

To introduce some complexity, we repeated the above experiments on the same two sentences, but the entire sentence had been lowercased this time. We did this to verify the model's sensitivity towards casing. As shown in the table above, the predictions made by each of the models on lowercased inputs varied significantly, and the models committed more entity misclassifications than before. However, the predictions made by the model augmented with knowledge remained unaltered and unaffected, which verified the applicability and adaptability of our proposed method to a real-world scenario where the raw text is not guaranteed to carry any specific formatting.

\section{Conclusion and Future Work}

This work proposed a novel world knowledge augmentation technique that leveraged large knowledge bases represented as fact triplets and successfully extracted relevant information for word-level augmentation. The model was trained and tested in an NER setting. Experimental results showed that knowledge level representation learning outperformed most NER systems in literature and made the model highly applicable to a real-world scenario by accurately predicting entities in random pieces of text.

Since we augmented features at the word level, we believe our method could facilitate many other NLP tasks, such as Chunking, Word Sense Disambiguation, Question Answering, etc. Therefore, as future work, we plan to test the applicability of the proposed methods on other NLP tasks as well. Our intuition says that any system can leverage the proposed system as a general knowledge representation learning tool. Moreover, being among the very few works in this direction, we see an ample scope of improvement. For instance, the Knowledge Graph Embedding model was trained separately on a Masked Language Modelling task, and then the trained model was used on the task at hand. This restricted the model from interacting and learning from the task at hand, NER in our case. We believe that a technique to incorporate and train the NER model with the knowledge representation module can be more beneficial. Another improvement that could be made lies in the entity shortlisting step. Although the technique is quite reliable, it does not consider any semantic information about the entities. Different entities in a knowledge base can be highly correlated to each other and yet have different names. Therefore, we plan to improve the entity shortlisting technique further for more accurate and robust shortlisting.

\bibliography{aaai22}

\begin{thebibliography}{58}
\providecommand{\natexlab}[1]{#1}

\bibitem[{Akbik, Blythe, and Vollgraf(2018)}]{akbik2018}
Akbik, A.; Blythe, D.; and Vollgraf, R. 2018.
\newblock Contextual String Embeddings for Sequence Labeling.
\newblock In Bender, E.~M.; Derczynski, L.; and Isabelle, P., eds.,
  \emph{Proceedings of the 27th International Conference on Computational
  Linguistics, {COLING} 2018, Santa Fe, New Mexico, USA, August 20-26, 2018},
  1638--1649. Association for Computational Linguistics.

\bibitem[{Annervaz, Chowdhury, and Dukkipati(2018)}]{annervaz2018learning}
Annervaz, K.~M.; Chowdhury, S. B.~R.; and Dukkipati, A. 2018.
\newblock Learning beyond datasets: Knowledge Graph Augmented Neural Networks
  for Natural language Processing.
\newblock arXiv:1802.05930.

\bibitem[{Antony and G~S(2015)}]{articleir_ner}
Antony, B.; and G~S, M. 2015.
\newblock Content-based information retrieval by named entity recognition and
  verb semantic role labelling.
\newblock 21: 1830--1848.

\bibitem[{Armour, Japkowicz, and Matwin(2005)}]{article_tc_ner}
Armour, Q.; Japkowicz, N.; and Matwin, S. 2005.
\newblock The Role of Named Entities in Text Classification.

\bibitem[{Babych and Hartley(2003)}]{babych-hartley-2003-improving}
Babych, B.; and Hartley, A. 2003.
\newblock Improving Machine Translation Quality with Automatic Named Entity
  Recognition.
\newblock In \emph{Proceedings of the 7th International {EAMT} workshop on {MT}
  and other language technology tools, Improving {MT} through other language
  technology tools, Resource and tools for building {MT} at {EACL} 2003}.

\bibitem[{Chiu and Nichols(2016)}]{chiu-nichols-2016-named}
Chiu, J.~P.; and Nichols, E. 2016.
\newblock Named Entity Recognition with Bidirectional {LSTM}-{CNN}s.
\newblock \emph{Transactions of the Association for Computational Linguistics},
  4: 357--370.

\bibitem[{Coates and Bollegala(2018)}]{coates-bollegala-2018-frustratingly}
Coates, J.; and Bollegala, D. 2018.
\newblock Frustratingly Easy Meta-Embedding {--} Computing Meta-Embeddings by
  Averaging Source Word Embeddings.
\newblock In \emph{Proceedings of the 2018 Conference of the North {A}merican
  Chapter of the Association for Computational Linguistics: Human Language
  Technologies, Volume 2 (Short Papers)}, 194--198. New Orleans, Louisiana:
  Association for Computational Linguistics.

\bibitem[{Dai et~al.(2019)Dai, Yang, Yang, Carbonell, Le, and
  Salakhutdinov}]{DBLP:journals/corr/abs-1901-02860}
Dai, Z.; Yang, Z.; Yang, Y.; Carbonell, J.~G.; Le, Q.~V.; and Salakhutdinov, R.
  2019.
\newblock Transformer-XL: Attentive Language Models Beyond a Fixed-Length
  Context.
\newblock \emph{CoRR}, abs/1901.02860.

\bibitem[{Das et~al.(2017)Das, Neelakantan, Belanger, and
  McCallum}]{das-etal-2017-chains}
Das, R.; Neelakantan, A.; Belanger, D.; and McCallum, A. 2017.
\newblock Chains of Reasoning over Entities, Relations, and Text using
  Recurrent Neural Networks.
\newblock In \emph{Proceedings of the 15th Conference of the {E}uropean Chapter
  of the Association for Computational Linguistics: Volume 1, Long Papers},
  132--141. Valencia, Spain: Association for Computational Linguistics.

\bibitem[{Denk and
  Peleteiro~Ramallo(2020)}]{denk-peleteiro-ramallo-2020-contextual}
Denk, T.~I.; and Peleteiro~Ramallo, A. 2020.
\newblock Contextual {BERT}: Conditioning the Language Model Using a Global
  State.
\newblock In \emph{Proceedings of the Graph-based Methods for Natural Language
  Processing (TextGraphs)}, 46--50. Barcelona, Spain (Online): Association for
  Computational Linguistics.

\bibitem[{Devlin et~al.(2018)Devlin, Chang, Lee, and Toutanova}]{bert}
Devlin, J.; Chang, M.-W.; Lee, K.; and Toutanova, K. 2018.
\newblock BERT: Pre-training of Deep Bidirectional Transformers for Language
  Understanding.

\bibitem[{dos Santos and Guimarães(2015)}]{santos2015boosting}
dos Santos, C.~N.; and Guimarães, V. 2015.
\newblock Boosting Named Entity Recognition with Neural Character Embeddings.
\newblock arXiv:1505.05008.

\bibitem[{Farouk(2018)}]{8639211}
Farouk, M. 2018.
\newblock Sentence Semantic Similarity based on Word Embedding and WordNet.
\newblock In \emph{2018 13th International Conference on Computer Engineering
  and Systems (ICCES)}, 33--37.

\bibitem[{Guo et~al.(2019)Guo, Qiu, Liu, Shao, Xue, and
  Zhang}]{guo2019startransformer}
Guo, Q.; Qiu, X.; Liu, P.; Shao, Y.; Xue, X.; and Zhang, Z. 2019.
\newblock Star-Transformer.
\newblock arXiv:1902.09113.

\bibitem[{He et~al.(2020)He, Wu, Yin, and Cai}]{qizhen2020}
He, Q.; Wu, L.; Yin, Y.; and Cai, H. 2020.
\newblock Knowledge-Graph Augmented Word Representations for Named Entity
  Recognition.
\newblock \emph{Proceedings of the AAAI Conference on Artificial Intelligence},
  34: 7919--7926.

\bibitem[{Hochreiter and Schmidhuber(1997)}]{articlelstm}
Hochreiter, S.; and Schmidhuber, J. 1997.
\newblock Long Short-term Memory.
\newblock \emph{Neural computation}, 9: 1735--80.

\bibitem[{Jiang, Wang, and Wang(2019)}]{jiang-etal-2019-adaptive}
Jiang, X.; Wang, Q.; and Wang, B. 2019.
\newblock Adaptive Convolution for Multi-Relational Learning.
\newblock In \emph{Proceedings of the 2019 Conference of the North {A}merican
  Chapter of the Association for Computational Linguistics: Human Language
  Technologies, Volume 1 (Long and Short Papers)}, 978--987. Minneapolis,
  Minnesota: Association for Computational Linguistics.

\bibitem[{Ke et~al.(2020)Ke, Ji, Liu, Zhu, and Huang}]{ke2020sentilare}
Ke, P.; Ji, H.; Liu, S.; Zhu, X.; and Huang, M. 2020.
\newblock SentiLARE: Sentiment-Aware Language Representation Learning with
  Linguistic Knowledge.
\newblock arXiv:1911.02493.

\bibitem[{Kim et~al.(2015)Kim, Song, Park, and Lee}]{7451560}
Kim, A.-Y.; Song, H.-J.; Park, S.-B.; and Lee, S.-J. 2015.
\newblock A re-ranking model for dependency parsing with knowledge graph
  embeddings.
\newblock In \emph{2015 International Conference on Asian Language Processing
  (IALP)}, 177--180.

\bibitem[{Kingma and Ba(2017)}]{kingma2017adam}
Kingma, D.~P.; and Ba, J. 2017.
\newblock Adam: A Method for Stochastic Optimization.
\newblock arXiv:1412.6980.

\bibitem[{Kouris, Alexandridis, and
  Stafylopatis(2019)}]{kouris-etal-2019-abstractive}
Kouris, P.; Alexandridis, G.; and Stafylopatis, A. 2019.
\newblock Abstractive Text Summarization Based on Deep Learning and Semantic
  Content Generalization.
\newblock In \emph{Proceedings of the 57th Annual Meeting of the Association
  for Computational Linguistics}, 5082--5092. Florence, Italy: Association for
  Computational Linguistics.

\bibitem[{Kuru, Can, and Yuret(2016)}]{kuru2016}
Kuru, O.; Can, O.~A.; and Yuret, D. 2016.
\newblock Charner: Character-level named entity recognition.
\newblock 911–921.

\bibitem[{Lafferty, McCallum, and Pereira(2001)}]{10.5555/645530.655813}
Lafferty, J.~D.; McCallum, A.; and Pereira, F. C.~N. 2001.
\newblock Conditional Random Fields: Probabilistic Models for Segmenting and
  Labeling Sequence Data.
\newblock In \emph{Proceedings of the Eighteenth International Conference on
  Machine Learning}, ICML '01, 282–289. San Francisco, CA, USA: Morgan
  Kaufmann Publishers Inc.
\newblock ISBN 1558607781.

\bibitem[{Lample et~al.(2016)Lample, Ballesteros, Subramanian, Kawakami, and
  Dyer}]{lample-etal-2016-neural}
Lample, G.; Ballesteros, M.; Subramanian, S.; Kawakami, K.; and Dyer, C. 2016.
\newblock Neural Architectures for Named Entity Recognition.
\newblock In \emph{Proceedings of the 2016 Conference of the North {A}merican
  Chapter of the Association for Computational Linguistics: Human Language
  Technologies}, 260--270. San Diego, California: Association for Computational
  Linguistics.

\bibitem[{Le et~al.(2018)Le, Postma, Urbani, and Vossen}]{le-etal-2018-deep}
Le, M.; Postma, M.; Urbani, J.; and Vossen, P. 2018.
\newblock A Deep Dive into Word Sense Disambiguation with {LSTM}.
\newblock In \emph{Proceedings of the 27th International Conference on
  Computational Linguistics}, 354--365. Santa Fe, New Mexico, USA: Association
  for Computational Linguistics.

\bibitem[{Li et~al.(2017)Li, Dong, Wang, Chou, and
  Ma}]{li-etal-2017-leveraging-linguistic}
Li, P.-H.; Dong, R.-P.; Wang, Y.-S.; Chou, J.-C.; and Ma, W.-Y. 2017.
\newblock Leveraging Linguistic Structures for Named Entity Recognition with
  Bidirectional Recursive Neural Networks.
\newblock In \emph{Proceedings of the 2017 Conference on Empirical Methods in
  Natural Language Processing}, 2664--2669. Copenhagen, Denmark: Association
  for Computational Linguistics.

\bibitem[{Li et~al.(2020{\natexlab{a}})Li, Sun, Meng, Liang, Wu, and
  Li}]{li2020dice}
Li, X.; Sun, X.; Meng, Y.; Liang, J.; Wu, F.; and Li, J. 2020{\natexlab{a}}.
\newblock Dice Loss for Data-imbalanced NLP Tasks.
\newblock arXiv:1911.02855.

\bibitem[{Li et~al.(2020{\natexlab{b}})Li, Li, Suhara, Doan, and Tan}]{Li_2020}
Li, Y.; Li, J.; Suhara, Y.; Doan, A.; and Tan, W.-C. 2020{\natexlab{b}}.
\newblock Deep entity matching with pre-trained language models.
\newblock \emph{Proceedings of the VLDB Endowment}, 14(1): 50–60.

\bibitem[{Lin et~al.(2002)Lin, Yang, Tseng, and
  Huang}]{The_Structure_of_Polysemy}
Lin, J.-Y.; Yang, C.-H.; Tseng, S.-C.; and Huang, C.-R. 2002.
\newblock The Structure of Polysemy: A study of multi-sense words based on
  WordNet.
\newblock In \emph{Language, Information, and Computation - Proceedings of the
  16th Pacific Asia Conference}, 320--329.

\bibitem[{Lin et~al.(2015)Lin, Liu, Luan, Sun, Rao, and
  Liu}]{lin-etal-2015-modeling}
Lin, Y.; Liu, Z.; Luan, H.; Sun, M.; Rao, S.; and Liu, S. 2015.
\newblock Modeling Relation Paths for Representation Learning of Knowledge
  Bases.
\newblock In \emph{Proceedings of the 2015 Conference on Empirical Methods in
  Natural Language Processing}, 705--714. Lisbon, Portugal: Association for
  Computational Linguistics.

\bibitem[{Lukovnikov et~al.(2017)Lukovnikov, Fischer, Lehmann, and
  Auer}]{10.1145/3038912.3052675}
Lukovnikov, D.; Fischer, A.; Lehmann, J.; and Auer, S. 2017.
\newblock Neural Network-Based Question Answering over Knowledge Graphs on Word
  and Character Level.
\newblock In \emph{Proceedings of the 26th International Conference on World
  Wide Web}, WWW '17, 1211–1220. Republic and Canton of Geneva, CHE:
  International World Wide Web Conferences Steering Committee.
\newblock ISBN 9781450349130.

\bibitem[{Luo, Xiao, and Zhao(2019)}]{luo2019hierarchical}
Luo, Y.; Xiao, F.; and Zhao, H. 2019.
\newblock Hierarchical Contextualized Representation for Named Entity
  Recognition.
\newblock arXiv:1911.02257.

\bibitem[{Luoma and Pyysalo(2020)}]{luoma2020exploring}
Luoma, J.; and Pyysalo, S. 2020.
\newblock Exploring Cross-sentence Contexts for Named Entity Recognition with
  BERT.
\newblock arXiv:2006.01563.

\bibitem[{Ma and Hovy(2016)}]{ma2016endtoend}
Ma, X.; and Hovy, E. 2016.
\newblock End-to-end Sequence Labeling via Bi-directional LSTM-CNNs-CRF.
\newblock arXiv:1603.01354.

\bibitem[{Nasr, Badr, and Joun(2002)}]{10.5555/646815.708603}
Nasr, G.~E.; Badr, E.~A.; and Joun, C. 2002.
\newblock Cross Entropy Error Function in Neural Networks: Forecasting Gasoline
  Demand.
\newblock In \emph{Proceedings of the Fifteenth International Florida
  Artificial Intelligence Research Society Conference}, 381–384. AAAI Press.
\newblock ISBN 157735141X.

\bibitem[{Pennington, Socher, and Manning(2014)}]{pennington-etal-2014-glove}
Pennington, J.; Socher, R.; and Manning, C. 2014.
\newblock {G}lo{V}e: Global Vectors for Word Representation.
\newblock In \emph{Proceedings of the 2014 Conference on Empirical Methods in
  Natural Language Processing ({EMNLP})}, 1532--1543. Doha, Qatar: Association
  for Computational Linguistics.

\bibitem[{Pradhan et~al.(2013)Pradhan, Moschitti, Xue, Ng, Bj{\"o}rkelund,
  Uryupina, Zhang, and Zhong}]{pradhan-etal-2013-towards}
Pradhan, S.; Moschitti, A.; Xue, N.; Ng, H.~T.; Bj{\"o}rkelund, A.; Uryupina,
  O.; Zhang, Y.; and Zhong, Z. 2013.
\newblock Towards Robust Linguistic Analysis using {O}nto{N}otes.
\newblock In \emph{Proceedings of the Seventeenth Conference on Computational
  Natural Language Learning}, 143--152. Sofia, Bulgaria: Association for
  Computational Linguistics.

\bibitem[{Qian et~al.(2018)Qian, Santus, Jin, Guo, and Barzilay}]{qian2018}
Qian, Y.; Santus, E.; Jin, Z.; Guo, J.; and Barzilay, R. 2018.
\newblock GraphIE: A Graph-Based Framework for Information Extraction.

\bibitem[{Reimers and Gurevych(2019)}]{reimers2019sentencebert}
Reimers, N.; and Gurevych, I. 2019.
\newblock Sentence-BERT: Sentence Embeddings using Siamese BERT-Networks.
\newblock arXiv:1908.10084.

\bibitem[{Sang and Meulder(2003)}]{DBLP:journals/corr/cs-CL-0306050}
Sang, E. F. T.~K.; and Meulder, F.~D. 2003.
\newblock Introduction to the CoNLL-2003 Shared Task: Language-Independent
  Named Entity Recognition.
\newblock \emph{CoRR}, cs.CL/0306050.

\bibitem[{Shaw, Uszkoreit, and Vaswani(2018)}]{shaw-etal-2018-self}
Shaw, P.; Uszkoreit, J.; and Vaswani, A. 2018.
\newblock Self-Attention with Relative Position Representations.
\newblock In \emph{Proceedings of the 2018 Conference of the North {A}merican
  Chapter of the Association for Computational Linguistics: Human Language
  Technologies, Volume 2 (Short Papers)}, 464--468. New Orleans, Louisiana:
  Association for Computational Linguistics.

\bibitem[{Vaswani et~al.(2017)Vaswani, Shazeer, Parmar, Uszkoreit, Jones,
  Gomez, Kaiser, and Polosukhin}]{vaswani2017attention}
Vaswani, A.; Shazeer, N.; Parmar, N.; Uszkoreit, J.; Jones, L.; Gomez, A.~N.;
  Kaiser, L.; and Polosukhin, I. 2017.
\newblock Attention Is All You Need.
\newblock arXiv:1706.03762.

\bibitem[{Wang et~al.(2018)Wang, Li, Wang, Zhang, Shen, Zhang, Henao, and
  Carin}]{wang2018joint}
Wang, G.; Li, C.; Wang, W.; Zhang, Y.; Shen, D.; Zhang, X.; Henao, R.; and
  Carin, L. 2018.
\newblock Joint Embedding of Words and Labels for Text Classification.
\newblock arXiv:1805.04174.

\bibitem[{Wang et~al.(2020{\natexlab{a}})Wang, Huang, Wang, Dai, Jiang, Liu,
  Lyu, Zhu, and Wu}]{wang2020coke}
Wang, Q.; Huang, P.; Wang, H.; Dai, S.; Jiang, W.; Liu, J.; Lyu, Y.; Zhu, Y.;
  and Wu, H. 2020{\natexlab{a}}.
\newblock CoKE: Contextualized Knowledge Graph Embedding.
\newblock arXiv:1911.02168.

\bibitem[{Wang et~al.(2020{\natexlab{b}})Wang, Jiang, Bach, Wang, Huang, Huang,
  and Tu}]{wang2020automated}
Wang, X.; Jiang, Y.; Bach, N.; Wang, T.; Huang, Z.; Huang, F.; and Tu, K.
  2020{\natexlab{b}}.
\newblock Automated Concatenation of Embeddings for Structured Prediction.
\newblock arXiv:2010.05006.

\bibitem[{Wang et~al.(2021)Wang, Jiang, Bach, Wang, Huang, Huang, and
  Tu}]{wang2021improving}
Wang, X.; Jiang, Y.; Bach, N.; Wang, T.; Huang, Z.; Huang, F.; and Tu, K. 2021.
\newblock Improving Named Entity Recognition by External Context Retrieving and
  Cooperative Learning.
\newblock arXiv:2105.03654.

\bibitem[{Wang et~al.(2019)Wang, Shang, Liu, Lu, Liu, and
  Han}]{wang2019crossweigh}
Wang, Z.; Shang, J.; Liu, L.; Lu, L.; Liu, J.; and Han, J. 2019.
\newblock CrossWeigh: Training Named Entity Tagger from Imperfect Annotations.
\newblock arXiv:1909.01441.

\bibitem[{Weston, Chopra, and Bordes(2015)}]{weston2015memory}
Weston, J.; Chopra, S.; and Bordes, A. 2015.
\newblock Memory Networks.
\newblock arXiv:1410.3916.

\bibitem[{Xiao, Huang, and Zhu(2016)}]{xiao2015}
Xiao, H.; Huang, M.; and Zhu, X. 2016.
\newblock {T}rans{G} : A Generative Model for Knowledge Graph Embedding.
\newblock In \emph{Proceedings of the 54th Annual Meeting of the Association
  for Computational Linguistics (Volume 1: Long Papers)}, 2316--2325. Berlin,
  Germany: Association for Computational Linguistics.

\bibitem[{Xin et~al.(2018)Xin, Hart, Mahajan, and Ruvini}]{intnet}
Xin, Y.; Hart, E.; Mahajan, V.; and Ruvini, J.-D. 2018.
\newblock Learning Better Internal Structure of Words for Sequence Labeling.

\bibitem[{Yamada et~al.(2020)Yamada, Asai, Shindo, Takeda, and
  Matsumoto}]{yamada2020luke}
Yamada, I.; Asai, A.; Shindo, H.; Takeda, H.; and Matsumoto, Y. 2020.
\newblock LUKE: Deep Contextualized Entity Representations with Entity-aware
  Self-attention.
\newblock arXiv:2010.01057.

\bibitem[{Yan et~al.(2019)Yan, Deng, Li, and Qiu}]{yan2019tener}
Yan, H.; Deng, B.; Li, X.; and Qiu, X. 2019.
\newblock TENER: Adapting Transformer Encoder for Named Entity Recognition.
\newblock arXiv:1911.04474.

\bibitem[{Yang, Zhang, and Dong(2017)}]{yang2017}
Yang, J.; Zhang, Y.; and Dong, F. 2017.
\newblock Neural Reranking for Named Entity Recognition.

\bibitem[{Yao et~al.(2015)Yao, Liu, Liu, Li, and Anwar}]{yao2015}
Yao, L.; Liu, H.; Liu, Y.; Li, X.; and Anwar, M. 2015.
\newblock Biomedical Named Entity Recognition based on Deep Neutral Network.
\newblock \emph{International Journal of Hybrid Information Technology}, 8:
  279--288.

\bibitem[{Yu, Bohnet, and Poesio(2020)}]{yu2020named}
Yu, J.; Bohnet, B.; and Poesio, M. 2020.
\newblock Named Entity Recognition as Dependency Parsing.
\newblock arXiv:2005.07150.

\bibitem[{Zhang, Liu, and Song(2018)}]{zhang2018}
Zhang, Y.; Liu, Q.; and Song, L. 2018.
\newblock Sentence-State LSTM for Text Representation.

\bibitem[{Zhong et~al.(2015)Zhong, Zhang, Wang, Wan, and Chen}]{zhong2015}
Zhong, H.; Zhang, J.; Wang, Z.; Wan, H.; and Chen, Z. 2015.
\newblock Aligning Knowledge and Text Embeddings by Entity Descriptions.
\newblock 267--272.

\bibitem[{Zhou et~al.(2017)Zhou, Zheng, Xu, Qi, Bao, and Xu}]{zhou2017}
Zhou, P.; Zheng, S.; Xu, J.; Qi, Z.; Bao, H.; and Xu, B. 2017.
\newblock Joint Extraction of Multiple Relations and Entities by Using a Hybrid
  Neural Network.
\newblock 135--146.
\newblock ISBN 978-3-319-69004-9.

\end{thebibliography}

\end{document}